# 基于改进条件生成扩散模型的新能源日前场景生成方法

王长刚 [1,2]，刘伟 [2]，曹宇 [1,2]，梁栋 [3]，李扬 [1,2]，莫静山 [1,2]

（1．现代电力系统仿真控制与绿色电能新技术教育部重点实验室(东北电力大学)，吉林省 吉林市 132012；2．东北电力大学电气工程学院，吉林省 吉林市 132012；3．吉林供电公司变电二次检修中心，吉林省 吉林市 132012）

## The Day-ahead Scenario Generation Method for New Energy Based on an Improved Conditional Generative Diffusion Model

WANG Changgang[1,2], LIU Wei[2], CAO Yu[1,2], LIANG Dong[3], LI Yang[1,2], MO Jingshan[1,2]

(1.Key Laboratory of Modern Power System Simulation and Control and Renewable Energy Technology, Ministry of Education, Northeast Electric Power University, Jilin 132012, Jilin Province, China;
2. School of Electrical Engineering, Northeast Electric Power University, Jilin 132012, Jilin Province, China;
3. Jilin Power Supply Company Substation Secondary Maintenance Center, Jilin 132012, Jilin Province, China)

**ABSTRACT:** In the context of the rising share of new energy generation, accurately generating new energy output scenarios is crucial for day-ahead power system scheduling. Deep learning-based scenario generation methods can address this need, but their black-box nature raises concerns about interpretability. To tackle this issue, this paper introduces a method for day-ahead new energy scenario generation based on an improved conditional generative diffusion model. This method is built on the theoretical framework of Markov chains and variational inference. It first transforms historical data into pure noise through a diffusion process, then uses conditional information to guide the denoising process, ultimately generating scenarios that satisfy the conditional distribution. Additionally, the noise table is improved to a cosine form, enhancing the quality of the generated scenarios. When applied to actual wind and solar output data, the results demonstrate that this method effectively generates new energy output scenarios with good adaptability.

**摘要**：在新能源发电占比不断提高的背景下，如何准确生成新能源出力场景对电力系统日前调度至关重要。基于深度学习的场景生成方法能够完成上述任务，但其黑盒性质使之存在可解释性差等问题。对此，该文提出一种基于改进条件生成扩散模型的新能源日前场景生成方法。该方法的构建基于马尔可夫链和变分推理的理论框架，先通过扩散过程将历史数据转化为纯噪声，再由条件信息引导去噪过程从而生成满足条件分布的场景。并改进噪声表为余弦形式，提高生成场景质量。最后，将方法应用于实际风光出力数据中，结果表明，该方法能够有效生成新能源出力场景，且适应性较好。

**关键词**：新能源；不确定性；场景生成；改进条件生成扩散模型；余弦噪声表

## 0 引言

随着"双碳"目标的提出，我国的能源结构逐渐的从化石能源向新能源转型[1]。近年来，我国可再生能源呈现出高比例跃升发展的势头[2]。截至2024年3月底，全国可再生能源装机达到15.85亿千瓦，同比增长26%，约占我国总装机的52.9%，其中，风电和光伏发电之和突破11亿千瓦[3]。在当前新能源发电占比不断提高的背景下，如何有效的描述新能源的不确定性对于电力系统的规划和稳定安全运行至关重要。

目前，通常使用场景分析法来描述风电光伏出力的波动性[4]。该方法基于历史数据、统计分析、建立模型等来生成一系列可能发生的场景。由于日前风光出力场景是电力系统进行日前优化调度的关键依据，通过生成符合实际出力特性的场景，可以更好地预测风电和光伏的输出，从而优化电力系统的调度计划，以提高电力系统调度对其不确定性的鲁棒性。因此，如何构建准确的场景集成为当前研究新能源不确定性的一个重要方向。

现如今，国内外学者已开展了新能源场景生成的大量研究，场景生成问题可以分为不确定性建模、时间序列建模以及抽样方法三类[5]。其中，构建风光出力不确定性模型是研究场景生成的基础，现有方法主要有物理方法[6]、统计学方法[7-10]等。文献[6]考虑到太阳辐射和温度是光伏出力的两个最大的影响因素,建立了基于不确定理论的太阳辐射值预

---

基金项目：国家自然科学青年基金项目(52307084)。

Project Supported by National Natural Science Foundation of China (52307084).

测模型。文献[7,8]使用正态分布，文献[9]使用 Beta 分布，文献[10]使用 Weibull 分布来对风光出力的概率分布进行假设，但这些方法仅通过改变概率分布中某些参数来拟合数据，难以描述风光出力的高维非线性特征。因此，一系列考虑时空相关性的方法被提出，常见针对时间序列建模方法有场景树法、马尔可夫链法等。文献[11]采用 Markov 链的方法，考虑误差状态的时间关联性，利用马尔可夫链模型来描述不确定单元出力误差在时间轴上的变化过程。文献[12]根据场景树的方法，构造风光出力的最优分位点，合理划分整个调度区间，在子区间内遍历全部场景，形成场景集。但是不确定性模型所包含的概率信息无法直接应用于电力系统优化调度问题，通常需要使用抽样方法获得初始场景集。常用的抽样方法主要有蒙特卡洛抽样和拉丁超立方抽样方法[13]。

随着人工智能技术的不断发展，基于深度学习的生成模型在新能源的场景生成方向中的到了广泛的应用。主要包括变分编码器(variational auto-encoder，VAE)和生成对抗网络(generative adversarial networks，GAN)两类。文献[14]提出了一种基于条件变分自动编码器的风电光伏出力随机场景生成方法，其可以根据 one-hot 编码生成指定标签下的场景。文献[15]首次将生成对抗网络用于新能源场景生成。通过生成器与判别器两个独立深度学习网络之间的相互博弈，来提高生成器生成样本的真实性。文献[16]提出了条件生成对抗网络，将日前预测值作为条件，学习满足条件噪声分布与场景集的映射关系，并且采用 Wasserstein 距离作为判别器损失函数，提高了场景生成质量。文献[17]提出了改进条件生成对抗网络，耦合卷积运算局部泛化能力与 Transformer 全局注意力机制，建立了适应于提取多种分布式电源特征的 CGAN 模型，可以可控的生成场景。

目前为止，尽管新能源场景生成技术有了很大的成果，但仍具有以下的问题：(1)传统的场景生成方法通常需要对数据进行先验假设，且大多数模型仅针对某一特征分析，在多变的实际环境中很难找到广泛适用的模型；(2)基于深度学习的方法使用数据驱动的方式通过神经网络自适应学习数据的内在关系，适应性较好，但大多为黑盒模型，潜在空间与生成场景之间的关系未知，可解释性较差。

针对上述问题，本文提出改进条件生成扩散模型的新能源场景生成方法。该方法将历史实测数据与日前预测数据分别作为真实样本与条件输入模型，首先通过扩散过程将输入的历史实测数据变为纯噪声，其次在通过条件信息引导去噪过程逐渐的去除噪声，以生成满足条件分布的数据。通过模型的训练，扩散模型可以学习到满足条件分布的预测噪声，从而可以生成蕴含条件的实际场景。对于该模型，使用基于马尔可夫链和变分推理的理论框架。在前向过程，通过定义的噪声表，可以明确每一步添加的噪声量，从而了解模型在每一步生成的不确定性。在反向过程，每一步的去噪基于上一步的结果，使得整个生成过程更加透明。且通过变分下界最大化来训练模型，增加了可解释性。此外，本文将线性噪声表改进为余弦噪声表，提高了模型的适应性。最后，本文以实际的光伏、风电出力数据进行算例仿真，验证了本文所提出的改进条件生成扩散模型的有效性和适应性。

## 1 生成扩散模型

生成扩散模型最早是由 Jonathan Ho 等人于 2020 年提出的一种深度学习方法[18]，如附录 A 图 A1 所示为生成扩散模型的理论框图，主要用于计算机视觉、图像生成等领域。生成扩散模型是一个参数化的马尔可夫链，它使用变分推理进行训练，以产生与输入数据分布规律相匹配的样本。该方法主要由前向(扩散)过程与反向(去噪)过程与训练过程三部分组成。前向过程就是在原始数据中不断的添加噪声，直到数据变为无规律的纯噪声。接着通过反向过程逐步的去除噪声以恢复历史数据。在通过训练过程，扩散模型学习到实际数据的概率分布，从而可以生成符合实际分布规律的场景。基于生成扩散模型的新能源日前场景生成方法理论框图如图 1 所示，以一天的光伏出力功率为例，令其为原始数据 $x_0$ 进行绘制。

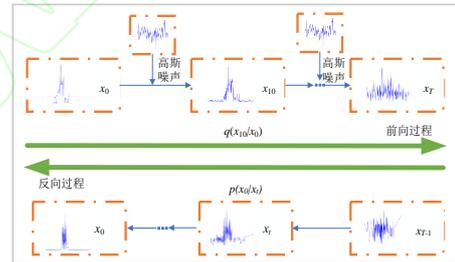

图 1 基于生成扩散模型的新能源日前场景生成方法理论框图

Fig.1 Theoretical framework diagram of the day-ahead new energy scenario generation method based on the generative diffusion model

若假设现有一组从高斯分布采样的噪声 $z$，生成扩散模型的最终目的就是对噪声 $z$ 进行变换，使其满足 $q(x_0)$ 的原始概率分布。其主要是通过扩散、去噪与训练过程完成，下面将详细阐述这三过程。

1) 前向过程

所谓前向过程，就是在每次迭代中，不断对原始数据 $x_0$ 添加高斯噪声，使其形成 $x_1, x_2, ..., x_T$，将这个过程进行参数化，可由下述公式表达：

$$q(x_{1:T} | x_0) := \prod_{t=1}^{T} q(x_t | x_{t-1}) \quad (1)$$

$$q(x_t | x_{t-1}) = N(x_t; \sqrt{1-\beta_t} x_{t-1}, \beta_t I) \quad (2)$$

式中：$T$ 为扩散总步骤。$N$ 为标准高斯分布函数。$\beta_t \epsilon (0,1)$ 为一个常数，控制着在每一步加入噪声的强度。并且 $\beta_0 < \beta_1 < ... < \beta_{t-1} < \beta_t < ... < \beta_T$，即 $\beta_t$ 随着 $t$ 增大也越来越大，加入噪声的程度也就越来越

大。$q(x_t|x_{t-1})$ 表示 $x_t$ 在 $x_{t-1}$ 下的分布属于正态分布，其平均值为 $\sqrt{1-\beta_t}x_{t-1}$，方差为 $\beta_t^2$。

对(3)式可以利用重参数化技巧，可以改写为：

$$x_t = \sqrt{1-\beta_t}x_{t-1} + \sqrt{\beta_t}\varepsilon_{t-1} \qquad (3)$$

由(3)式可以直观的看出噪声的添加方式。并且，当前状态 $x_t$ 的概率分布仅依赖于上一个状态 $x_{t-1}$，这个性质被称为马尔可夫性质。因此，整个扩散过程可以看做是一个马尔可夫链。所以，若给出原始数据 $x_0$，就可以通过计算得到加噪任意 $t$ 次之后的 $x_t$。

$$\begin{cases} \alpha_t = 1-\beta_t \\ \overline{\alpha}_t = \prod_{t=1}^{T}(1-\beta_t) \end{cases} \qquad (4)$$

$$x_t = \sqrt{\overline{\alpha}_t}x_0 + \sqrt{1-\overline{\alpha}_t}\varepsilon_t \qquad (5)$$

式中：$\varepsilon \epsilon N(0,I)$。

根据式(5)，就可通过一次采样获得状态 $t$ 时刻的样本。当 $T \to \infty$ 时，$x_T$ 的概率分布符合标准高斯噪声分布 $N(0,I)$。

在前向过程中，依据马尔可夫链的性质，每一步的结果仅与上一步有关，且通过定义的噪声表，独立的添加高斯噪声，从而可明确每一步添加的噪声量，此外每一步均为高斯分布，因此可以很容易的通过可视化展现每一步的效果，观察数据逐渐被噪声覆盖的过程，从而了解模型在每一步生成的不确定性，增加了模型的可解释性。

2) 反向过程

对于反向过程，也叫作逆扩散过程，就是通过一个满足噪声分布的 $x_T$ 来逐步的预测原始分布 $x_0$。反向过程仍然是一个马尔可夫链过程，所以其就是通过输入的 $x_t$ 来求 $x_{t-1}$ 的分布，即 $q(x_{t-1}|x_t)$。由贝叶斯公式可以得到：

$$q(x_{t-1}|x_t) = q(x_t|x_{t-1})\frac{q(x_{t-1})}{q(x_t)} \qquad (6)$$

对于式(6)，$q(x_t|x_{t-1})$ 可以由前向过程直接求取出来，但 $q(x_{t-1})$ 与 $q(x_t)$ 是未知的，但可以加入一个先决条件 $q(x_0)$，即：

$$q(x_{t-1}|x_t,x_0) = q(x_t|x_{t-1},x_0)\frac{q(x_{t-1}|x_0)}{q(x_t|x_0)} \qquad (7)$$

对于式(7)可以利用重参数技巧将概率分布运算转化为指数运算，从而得到：

$$q(x_{t-1}|x_t,x_0) \propto exp(-\frac{1}{2}[(\frac{\alpha_t}{\beta_t}+\frac{1}{1-\overline{\alpha}_{t-1}})x_{t-1}^2 - \\ (\frac{2\sqrt{\alpha_t}}{\beta_t}x_t + \frac{2\sqrt{\overline{\alpha}_{t-1}}}{1-\overline{\alpha}_{t-1}}x_0)x_{t-1} - C(x_t,x_0)] \qquad (8)$$

根据式(8)，可以得到 $x_{t-1}$ 的均值为 $x_t$ 与 $x_0$ 的函数，令其为 $\overline{\mu}_t(x_t,x_0)$，即：

$$\overline{\mu}_t(x_t,x_0) = \frac{\sqrt{\alpha_t}(1-\overline{\alpha}_{t-1})}{1-\overline{\alpha}_t}x_t + \frac{\sqrt{\overline{\alpha}_{t-1}}\beta_t}{1-\overline{\alpha}_t}x_0 \qquad (9)$$

通过式(5)，式(8)，式(9)可以得到 $x_{t-1}$ 的方差与均值分别为：

$$\overline{\beta}_t = \frac{1-\overline{\alpha}_{t-1}}{1-\overline{\alpha}_t} \cdot \beta_t \qquad (10)$$

$$\overline{\mu}_t = \frac{1}{\sqrt{\alpha_t}}(x_t - \frac{1-\alpha_t}{\sqrt{1-\overline{\alpha}_t}}\varepsilon_t) \qquad (11)$$

由上式可以看出，后验高斯分布的均值只与超参数，$x_t$，$\varepsilon_t$ 有关。方差只与超参数有关。

根据上述结果，可以给出生成扩散模型的反向过程的去噪方式为：

$$p_\theta(x_{0:T}) := p(x_T)\prod_{t=1}^{T}p_\theta(x_{t-1}|x_t) \qquad (12)$$

$$p_\theta(x_{t-1}|x_t) := N(x_{t-1};\mu_\theta(x_t,t),\varepsilon_\theta(x_t,t)) \qquad (13)$$

式中：$p_\theta(x_{0:T})$ 表示近似 $q(x_{0:T})$ 的概率分布。$\mu_\theta(x_t,t)$ 表示后验高斯分布的均值，$\varepsilon_\theta(x_t,t)$ 表示 $t$ 时刻添加的噪声，是一个未知可学习的参数。

在反向过程中，与前向过程类似，依据马尔可夫链的性质，可以对去噪过程中的每一步进行可视化，使得数据如何从噪声逐步恢复到原始数据的过程变得直观和透明。

3) 训练过程

对于训练过程，通过极大似然估计，来找到反向过程中马尔可夫链变换的概率分布。即最小化负对数似然：

$$L = \mathrm{E}_q[-\log p_\theta(x_0)] \qquad (14)$$

式中：E 表示对应分布的期望值。这个过程类似变分编码过程，因此可以使用变分下界(Variable Lower Boundary, VLB)来优化负对数似然，而变分下界在数学上有着明确的可解释性。

KL 散度是一种不对称距离度量，主要用于比较两个概率分布的差异程度，即：

$$L_t = D_{KL}(q(x_{t-1}|x_t), p_\theta(x_{t-1}|x_t)) \qquad (15)$$

$$\mathrm{E}_{q(x_0)}p_\theta(x_0) \leq L_{\mathrm{VLB}} = \mathrm{E}_{q(x_{1:T}|x_0)}\left[\log\frac{q(x_{1:T}|x_0)}{p_\theta(x_{T:0})}\right] \qquad (16)$$

式中：$L_{\mathrm{VLB}}$ 表示变分下界。进一步可以写出交叉熵的上界，并化简可以得到：

$$L_{\mathrm{VLB}} = \mathrm{E}_q[\underbrace{D_{KL}(q(x_T|x_0) \| p_\theta(x_T))}_{L_T} + \\ \sum_{t=2}^{T}\underbrace{D_{KL}(q(x_{t-1}|x_t,x_0) \| p_\theta(x_{t-1}|x_t))}_{L_{t-1}} \qquad (17) \\ \underbrace{-\log p_\theta(x_0|x_1)]}_{L_0}$$

式中：$L_T$ 表达为前向过程,因为没有可学习的参数，

并且 $x_T$ 为纯噪声，因此 $L_T$ 可以视为常量，可以忽略。$L_0$ 与 $L_{t-1}$ 表示为可优化的项，对于 $L_0$ 的优化效果远比不上 $L_{t-1}$，因此模型只对 $L_{t-1}$ 进行优化。$L_{t-1}$ 表示了 $q(x_{t-1}|x_t,x_0)$ 与 $p_\theta(x_{t-1}|x_t)$ 两个分布的 KL 散度无限逼近。对于前者属于高斯分布，其方差与均值分别为式(10)(11)所示。对于后者也属于高斯分布，是神经网络所期望拟合的目标分布，其均值需要估计，而方差被设置为与 $\beta_t$ 有关的固定值。因此，$L_{t-1}$ 的优化目标就是这两个分布均值的二范数，即：

$$L_{t-1} = E_q[\|\overline{\mu_t}(x_t,x_0) - \mu_\theta(x_t,t)\|^2]$$
$$= E_{x_0,\varepsilon}[\|\frac{1}{\sqrt{\alpha_t}}(x_t(x_0,\varepsilon) - \frac{\beta_t}{\sqrt{1-\overline{\alpha}_t}}\varepsilon) - \mu_\theta(x_t(x_0,\varepsilon),t)\|^2]$$
(18)

由上式可知，神经网络的优化目标就是使 $\mu_\theta(x_t,t)$ 近似 $\overline{\mu_t}(x_t,x_0)$，对于 $\mu_\theta$，只有噪声 $\varepsilon_\theta(x_t,t)$ 未知，因此可以通过网络预测出噪声 $\varepsilon_\theta$，在通过计算得到均值。所以，最终训练过程的目标函数为：

$$L_t = E_{x_0,\varepsilon}\left[\|\varepsilon - \varepsilon_\theta(\sqrt{\overline{\alpha}_t}x_0 + \sqrt{1-\overline{\alpha}_t}\varepsilon, t)\|^2\right] \quad (19)$$

式中：$\|\cdot\|$ 表示 L2 范数，$\varepsilon_\theta$ 表示预测的噪声。

## 2 基于改进条件生成扩散模型的新能源场景生成方法

### 2.1 条件生成扩散模型

根据国家风电和光伏电站的并网标准，电站需要在电力系统调度机构规定时间内，提交次日的发电功率预测曲线[19,20]。因此，新能源的日前场景生成问题，可以被视为在已知日前预测信息的基础上，构建次日新能源出力分布。由此，本文提出了改进条件生成扩散模型(Improve Condition Generative Diffusion Model, ICGDM)的新能源的日前场景生成方法。通过将新能源预测信息作为条件参与模型训练，并在反向过程引导新能源出力场景的生成。

在无条件情况下，反向过程 $p_\theta(x_{0:T})$ 用于定义最终的数据模型 $p_\theta(x_0)$。那么，在考虑条件下，公式(12)(13)则扩展为：

$$p_\theta(x_{0:T}|c) \coloneqq p(x_T)\prod_{t=1}^{T}p_\theta(x_{t-1}|x_t,c) \quad (20)$$

$$p_\theta(x_{t-1}|x_t,c) \coloneqq N(x_{t-1};\mu_\theta(x_t,t|c),\varepsilon_\theta(x_t,t|c)) \quad (21)$$

式中：$c$ 表示条件信息。由于公式(21)与公式(13)具有相同的参数化，且两者之间的差异只是 $\varepsilon_\theta$ 的形式，因此可以使用无条件模型训练程序。但此时神经网络所预测的目标变为满足条件分布下的噪声 $\varepsilon_\theta(x_t,t,c)$，则此时损失函数为：

$$L_t = E_{x_0,\varepsilon}\left[\|\varepsilon(x_t,t) - \varepsilon_\theta(x_t,t,c)\|^2\right] \quad (22)$$

上述公式的推导主要参考文献[21]。ICGDM 在经过模型训练后，这时采样的噪声 $\varepsilon_\theta$ 中就蕴含了条件信息，从而引导反向过程向着满足条件分布的方向去噪，从而生成满足条件分布的场景。

### 2.2 改进条件生成扩散模型

由于在前向过程使用超参数 $\beta_t$ 来控制加入噪声强度。在原方法中，$\beta_t$ 线性变化，如附录 A 图 A2(a)所示。但这种线性添加噪声的方式对于某些样本并不适用，如图2所示，以一天风电数据为原始样本，共进行扩散 100 步，每 10 步进行采样。在使用线性噪声表的方式添加噪声时，扩散后 1/3 部分基本全为纯噪声，这对后向采样生成数据是不利的。针对该问题，本文采用余弦噪声表的方式添加噪声，如附录 A 图 A2(b)所示。

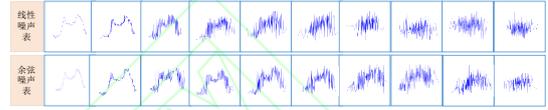

图 2 不同加噪方式对比图
**Fig.2 Comparison diagram of different noise adding modes**

首先，为了方便计算，定义 $\overline{\alpha}_t$ 为：

$$\begin{cases}\overline{\alpha}_t = \dfrac{f(t)}{f(0)} \\ f(t) = \cos(\dfrac{t/T+s}{1+s} \cdot \dfrac{\pi}{2})^2\end{cases} \quad (23)$$

从而，可以得到 $\beta_t$ 为：

$$\beta_t = 1 - \frac{\overline{\alpha}_t}{\overline{\alpha}_{t-1}} \quad (24)$$

此外，为了防止在扩散过程中当 $t=T$ 时出现奇点，令 $\beta_t < 0.999$。而且使用一个小的偏移量 $s$，防止 $\beta_t$ 在 $t=0$ 时太小，增加了神经网络预测噪声的准确性。

在使用余弦噪声表时，$\overline{\alpha}_t$ 在中间的变化过程中有一个线性下降。而在当 $t=0$ 与 $t=T$ 时的极值附近变化很小，防止了噪声的突然变化。如图3所示，为 $\overline{\alpha}_t$ 的变化对比图，图中纵轴代表 $\overline{\alpha}_t$ 的变化趋势，横轴代表扩散步数 $T$。从图中可以看出，使用线性噪声表 $\overline{\alpha}_t$ 更快的趋近于零，可以更快的破坏原始信息，使样本变为纯噪声。此外由图2更加直观了表示了使用余弦噪声表添加噪声的方式。

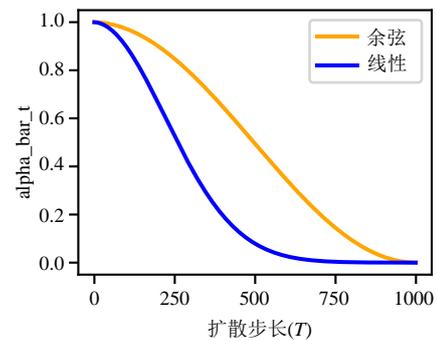

图 3 $\overline{\alpha}_t$ 的变化对比图
**Fig.3 Comparison of changes in $\overline{\alpha}_t$**

## 2.3 神经网络结构设计

在原方法中，主要使用 U-shaped Network(UNet)网络来预测所添加的噪声 $\varepsilon_\theta(x_t,t)$，但是 UNet 网络主要适用于生物医学图像处理领域，对于新能源场景方面已然不是最优方法。因此，本文根据 UNet 网络，结合文献[22]设计神经网络结构如附录 A 图 A3 所示。

首先，该神经网络结构基于膨胀卷积设计。膨胀卷积就是在跳过一些输入值的情况下，使卷积滤波器可以覆盖更大的输入区间。卷积核可以根据膨胀因子均匀的采样输入数据。这个过程类似 UNet 网络中的池化层，但与之不同的是，本方法输出的大小维度与输入相同。使用膨胀卷积的主要优点是可以明显的增加感受野，即在相同的计算成本下可以获取更大范围的信息。

其次，整个神经网络由 $N$ 个残差层组成。每个残差层包含一个残差连接与跳跃连接。其可以防止在训练深层网络时出现梯度消失或爆炸等，并且，这些连接还可以加速网络的收敛。在该模型中，令每一个残差层中的膨胀因子翻倍，即 $\{1,2,4,...,2^{N-1}\}$。因此随着深度的增加，可以成指数级的增加膨胀卷积的感受野，且网络的整体容量也随之增加。

接着，为了更好的挖掘历史数据之间的关系，使用门控激活单元作为非线性激活形式。该激活单元结合 Sigmoid 激活函数和双曲正切（tanh）激活函数，远比修正线性激活函数(ReLU)的效果好。门激活单元的公式如下：

$$\tanh(W_{f,k} * x) \odot \sigma(W_{g,k} * x) \tag{25}$$

式中：$W$ 是可学习的卷积滤波器，$f$ 和 $g$ 分别表示滤波器和门，$\sigma(\cdot)$ 表示 Sigmoid 算子，$k$ 为残差层索引。在每个残差层中，都会有一个门控激活单元。

然后，所预测噪声 $\varepsilon_\theta(x_t,t)$ 有个最大的特点，那就是他有两个输入。第一个是样本 $x_t$，第二个是时间步 $t$。因为 $\varepsilon_\theta(x_t,t)$ 必须针对不同的步数产生不同的输出，因此在模型中嵌入时间步 $t$ 是非常重要的。本文将时间步 $t$ 转换为包含不同频率的正弦和余弦向量，使用维度为 64 的向量来嵌入时间步 $t$，其公式如下：

$$t = \begin{bmatrix} \sin\left(10^{\frac{0\times 4}{31}} t\right),\cdots,\sin\left(10^{\frac{31\times 4}{31}} t\right), \\ \cos\left(10^{\frac{0\times 4}{31}} t\right),\cdots,\cos\left(10^{\frac{31\times 4}{31}} t\right) \end{bmatrix} \tag{26}$$

由附录A图A4可以看出，在生成嵌入后，首先经过两个全连接层，这两个层被所有残差层共享，因此参数也共享。接着，会再次通过一个全连接层，将前两层的输出映射到一个嵌入向量上，以作为每一个残差层时间步的输入。

最后，为条件信息的嵌入。条件信息由输入特征向量形成，与时间步 $t$ 的嵌入类似，条件信息将经过三层结构嵌入到网络中。前两层为所有残差共享的全连接层，如图A4中条件块正是条件信息经过这两个通用全连接层得到的输出。最后在通过一层一维卷积使条件信息嵌入网络，从而使条件信息有效影响神经网络的输出。该神经网络的具体参数如附录A表A1所示。

## 2.4 基于改进生成扩散模型的新能源场景生成方法框架

基于改进条件生成扩散模型的新能源场景生成方法框架如图 4 所示。

步骤 1：将光伏、风电的历史日前预测数据与实测数据作为该方法的输入。对原始数据进行预处理，将其归一化至[0,1]。

步骤 2：通过改进条件生成扩散模型进行加噪处理。对于原始样本实测数据 $x_0$ 使用余弦噪声表的方式，通过式(5)对其依次加入满足标准高斯分布的噪声。直到满足设置的扩散步数，使原始分布变为纯噪声。

步骤 3：为训练过程。将加入噪声的样本 $x_t$、位置编码 $t$ 与日前预测数据 $c$ 输入到神经网络中，神经网络根据 $t$ 生成正弦和余弦位置编码将三者进行结合，从而预测 $x_t$ 中所加入的噪声。将神经网络预测的满足条件分布的噪声与之前所采样的随机噪声求 L2 损失函数，计算梯度，根据 Adam 优化算法更新权重。重复这个过程，直至网络收敛，神经网络训练完成。

步骤 4：为新能源场景生成过程。从标准正态分布中采样出 $x_T$，然后根据神经网络所预测的满足条件分布的噪声，计算出样本噪声的均值，利用重参数技巧，得到 $x_{t-1}$。通过依次迭代，求解出满足条件分布的场景 $x_0$。

## 2.5 生成场景评价指标

为了保证生成场景的质量，场景生成应该最大程度的拟合新能源不确定性的特性。本文构建关于生成场景的有效性、准确性、随机性的评价体系，采用自相关系数、覆盖率、功率区间宽度、欧式距离平均值作为为评价指标。

1)有效性

由于光照、风速等因素的影响，新能源出力应有较强的时间相关性，因此分析时序特征是验证日前生成场景有效性的关键[23]。本文选择自相关系数对生成场景进行分析验证。自相关系数反应了不同滞后时间 $\tau$ 的关联程度。随着滞后时间的增加，自相关系数应逐渐减小。因此可以通过自相关系数的大小及变化特点来衡量生成场景的有效性。

自相关系数(Autocorrelation Function, ACF)公式如下所示。

$$R(\tau) = \frac{E[(S_t - \mu)(S_{t+\tau} - \mu)]}{\sigma^2} \tag{27}$$

式中：$S_t$ 表示真实场景或生成场景在 $t$ 时刻出力的值；$\tau$ 表示时间间隔；$\mu$ 表示风光出力的均值；$\sigma^2$ 表示风光出力的方差。

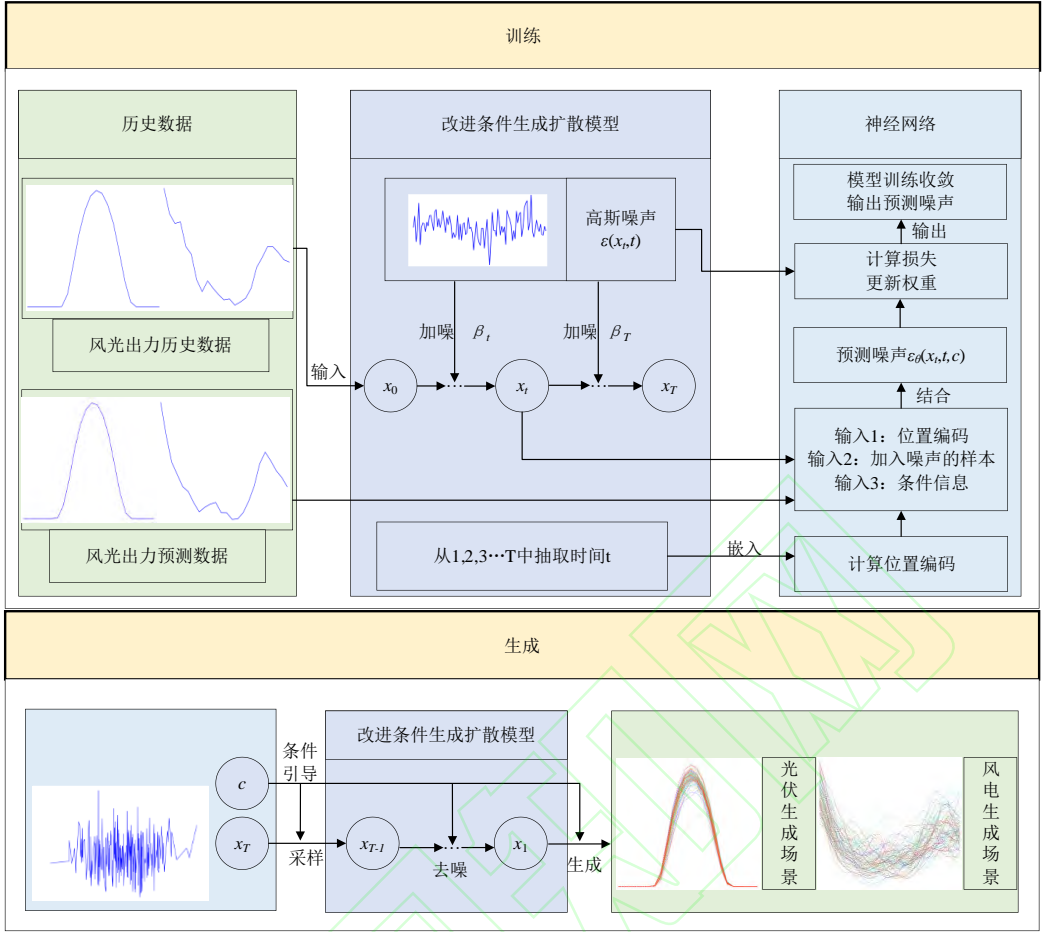

图 4 基于改进条件生成扩散模型的新能源场景生成方法框架
Fig.4 New energy scenario generation method framework based on improved condition generation diffusion model

2)准确性

本文选择覆盖率与功率区间宽度来评价生成场景的准确性[24]。覆盖率用于衡量生成场景的置信区间对实际数据点的包含程度。其值越大，表明实际场景落入生成场景范围的概率越大。功率区间宽度用于评价生成场景聚集不确定信息的表达能力。其值越小，表明生成场景越稳定。但覆盖率与功率区间宽度均不能单独作为评价指标，应该综合比较。当覆盖率越大功率区间越小时，生成的场景越能准确描述新能源的不确定性。当覆盖率相似功率区间越小，模型的精确性越好。

覆盖率(coverage rate，CR)和功率区间宽度(power interval width，PIW)的公式如下所示。

$$\begin{cases} \delta_{CR_\theta} = \left[\frac{1}{N}\sum_{n=1}^{N} 1(P_{n,\theta}^{\min} \leq P_n \leq P_{n,\theta}^{\max})\right] \times 100\% \\ P_{n,\theta}^{\min} = \min(\omega_{w,n}), w=1,...,W \\ P_{n,\theta}^{\max} = \max(\omega_{w,n}), w=1,...,W \end{cases} \quad (28)$$

式中：$\delta_{CR_\theta}$ 表示置信度 θ 下的覆盖率值；$N$ 表示测试集采样点总数；$1(\bullet)$ 为一个二进制变量，若括号里的条件满足，则取值为 1，否则为 0；$P_n$ 为第 $n$ 个采样点的功率；$P_{n,\theta}^{\min}$ 与 $P_{n,\theta}^{\max}$ 分别表示第 $n$ 个采样点在置信度 θ 下功率的最小值与最大值。

$$\delta_{PIW_\theta} = \frac{1}{N}\sum_{n=1}^{N}(P_{n,\theta}^{\max} - P_{n,\theta}^{\min}) \quad (29)$$

式中：$\delta_{PIW_\theta}$ 为置信度在 θ 下的功率区间平均宽度；$P_{n,\theta}^{\max} - P_{n,\theta}^{\min}$ 表示第 $n$ 个采样点在置信度 θ 下的功率区间宽度。

3)随机性

本文选择欧式距离平均值来评价生成场景的随机性[25]。欧式距离平均值反应了生成场景与真实场景的差异程度。其值越小，生成的场景越接近于实际值，模型越精确。

欧式距离平均值(Average Euclidean Distance, AED)的公式如下所示。

$$e_{AED} = \frac{1}{N}\sum_{n=1}^{N}\|R_n - R\|_2 \quad (30)$$

式中：$e_{AED}$ 表示生成场景与实际场景的欧式距离平均值；$R_n$ 表示生成的第 $n$ 个场景序列；$R$ 为历史实际场景序列；$\|\bullet\|_2$ 表示两个序列差的 L2 范数。

## 3 算例分析

### 3.1 算例概况

本文选择比利时地区全年光伏[26]与风电[27]的日前预测与实测数据作为算例数据。数据间隔为 1h，

风电光伏均有 8760 组数据。算例以一天 24 组的连续数据作为 1 天的样本，选取 315 天为训练集以训练模型，剩余 50 天为测试集验证模型的有效性。

改进条件生成扩散模型由深度学习框架Pytorch 搭建[28]。计算机硬件配置为：Intel(R) Core(TM) i5-1035G1 CPU 1.19 GHz，内存为 16.0GB，GPU 为 NVIDIA GeForce MX350，显存为 2.0GB。算例由本文所提出方法有效性验证，和本文方法与生成扩散模型、采用 Wasserstein 距离生成对抗网络场景生成方法效果进行仿真对比以及在不同数据集下的适应性三部分构成。

### 3.2 模型训练

采用 Adam 梯度下降算法进行模型训练。使用余弦噪声表添加噪声，超参数 $\beta_t$ 设置为从 0.0001 到 0.05。算例训练了两套模型分别用于光伏风电的场景生成，光伏的扩散步长为 250，风电的扩散步长为 200。其训练过程的损失量如图 5 所示，图中纵轴代表由式(19)所计算的损失值，横轴代表模型训练的轮次。

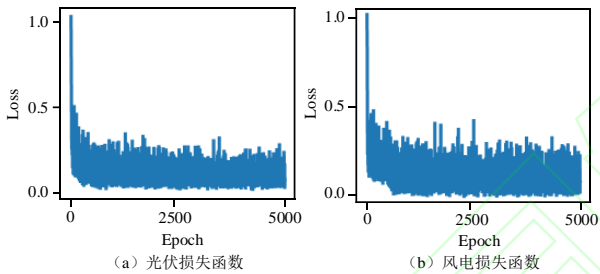

图 5 光伏和风电训练过程的损失量
**Fig.5 Amount of losses during pv and wind power training**

对于模型训练的起始阶段，神经网络并不能精确的预测所添加的噪声，因此风光的损失函数均会很大，随着训练次数的增加将逐渐下降并趋向稳定，此时模型收敛。

### 3.3 日前场景生成

模型训练完成后，从测试集中选取一天的光伏风电预测与实测数据作为测试样本。由于两者数据均来自于测试集，未经过模型训练，可以在一定程度上验证模型的泛化性。提取 100 组纯噪声，利用神经网络所预测前向过程加入的噪声，并使用预测信息进行反向过程的引导，根据改进条件生成扩散模型，分别生成光伏与风电的日前场景集，如图 6 所示。

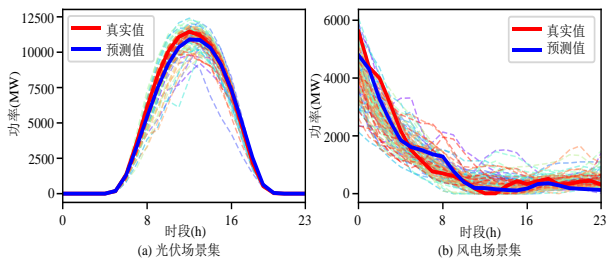

图 6 生成场景集
**Fig.6 Generated scene set**

图 6 中的虚线为基于 ICGDM 方法所生成的 100 个日前场景，红色的实线代表历史数据的真实值，蓝色的实线代表历史数据的预测值。由图 6 中可以看出，所生成的光伏、风电场景出力趋势与预测值基本相同，真实值能够被较好的包括在生成场景集中，且无明显的随机波动。由此证明了本文所提出的方法可以较好的描述新能源的出力趋势。

为了验证本文生成场景的有效性，本文计算了生成场景集与历史真实数据时间间隔在 0-6h 的自相关系数，如图 7 所示。图中红色圆点表示历史真实值的自相关系数，箱型图表示生成场景集的自相关系数分布图。

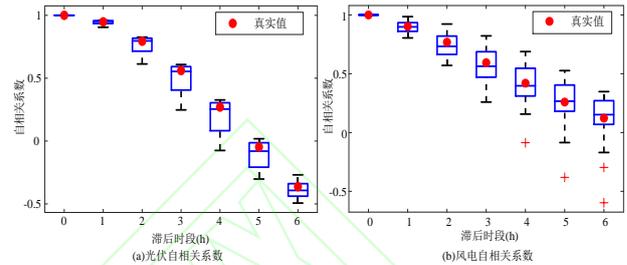

图 7 自相关系数箱型图
**Fig.7 Box diagram of autocorrelation coefficient**

从图 7 中可以看出，光伏风电的生成场景集均有一定的自相关性，并且随着滞后时间的增加自相关系数会逐渐的降低。其中，光伏的生成场景集自相关系数较集中，异常值较少，其中位线与历史真实值相距较近，符合光伏出力的变化特征。风电的生成场景集自相关系数更为均匀分布，其历史真实值在中位线附近。综上，光伏与风电所生成场景集的自相关系数均能把历史真实值包含在内，并且其变化趋势与真实值的变化趋势基本相同，符合新能源的变化特征，验证了本文方法的有效性。

### 3.4 方法对比

为了研究本文所提方法的优越性，本文设置了三种方案进行准确性与随机性的对比分析。

方案一为生成对抗网络的场景生成方法，通过生成器与判别器之间的博弈以学习到新能源场景的概率映射，并且使用 Wasserstein 距离作为判别器的损失函数，以增加模型的训练精度[16]。方案二为原始的生成扩散模型的场景生成方法，使用线性噪声表的方式添加噪声。方案三为本文所提出的改进条件生成扩散模型的新能源场景生成方法，使用余弦噪声表的方式添加噪声。

此外，由于新能源的出力具有显著的季节性特征，因此本文分别考虑季节性更为明显的夏季典型日与冬季典型日进行方法对比。

#### 3.4.1 夏季典型日

1)准确性

为了更直观的对比不同方案生成场景集的准确性，算例分别给出三种方案夏季典型日光伏与风电生成场景的置信区间如附录 A 图 A4、图 A5 和图 8 所示，以及为了更精确的对比三种方案，将上述结果整理成表，如附录 A 表 A2 所示。

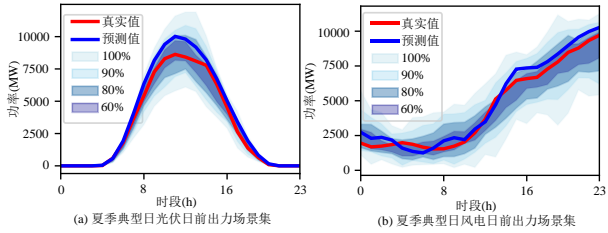

图 8 夏季典型日方案三生成场景集的置信区间
Fig.8 The confidence interval for the scenario set generated by the third typical summer day plan

由图 8 可以看出，方案三所生成的夏季典型日日前场景集有着较高的覆盖率，对于 100%置信区间均能完整的包含光伏风电的历史真实值，且生成场景集的变化趋势与日前预测值的变化趋势一致。

通过与附录 A 图 A4、图 A5 以及附录 A 表 A2 的对比，三种方案对于光伏风电生成场景集均有着较高的覆盖率，但是可以看出采用方案一的方法生成的场景集在不同时刻均有着较宽的功率区间，而方案二较方案一的情况功率区间宽度有所降低，但随着置信度的下降，方案一与方案二的覆盖率也有着明显的降低。而方案三不仅有着较高的覆盖率，其功率区间宽度也较窄。因此，使用方案三生成夏季典型日的光伏风电日前出力场景集更贴近新能源的变化特征。

2)随机性

统计三种方案夏季典型日生成场景集与历史真实值的欧氏距离平均值，结果如图 9 所示。

图 9 夏季典型日三种方案欧式距离平均值对比
Fig.9 Comparison of the average Euclidean distance among the three typical summer day plans

由图 9 可以看出，对于夏季典型日生成场景集的欧氏距离平均值，无论是波动性较小的光伏出力，还是波动性较大的风电出力，方案三的欧式距离平均值均为最小，其生成场景在距离上更接近实际场景。通过三种方案夏季典型日欧式距离平均值的对比，证明了本文提出的方法在刻画新能源场景生成随机性方面的优越性。

### 3.4.2 冬季典型日

1)准确性

如附录 A 图 A8 所示，为方案三生成冬季典型日的日前场景生成集。其中，光伏出力的季节性特征更为明显。在夏季典型日中，光伏出力大约在 4:00-20:00 时间段；在冬季典型日中，大约在 7:00-17:00 时间段，光照时间有着明显的缩短。对于风电而言，则有着较大的波动性，其出力基本无规律。但是，由方案三所生成冬季典型日的生成场景集，在 100%与 90%置信度下均能完整的包含风电光伏出力的历史真实值，且生成场景集的变化趋势与日前预测值的变化趋势基本相同。

通过与附录 A 图 A6、图 A7 以及附录 A 表 A3 三种方案的对比，与夏季典型日类似，方案三由于引入日前预测信息，且通过余弦噪声表添加噪声，因此在不同置信度下风电光伏生成场景集均有着较高的覆盖率和较窄的功率区间宽度。因此，使用方案三生成夏季典型日的光伏风电日前出力场景集更贴近新能源的变化特征。

2)随机性

如附录 A 表 A4 所示，为三种方案下冬季典型日生成场景集与历史真实值的欧式距离平均值。从表中可以看出，方案三的欧式距离平均值在光伏与风电上均小于方案一与方案二，由此说明了本文所提方法在生成冬季典型日的场景上其距离更接近于实际场景，随机性较小。

### 3.5 适用性分析

为了验证本文所提方法的适用性，算例以波动性更大的风电为例，分别使用瓦隆和弗兰德伦风电场[27]的全年日前预测及实测数据作为原始数据训练模型。从训练集中随机选取一天，使用三种方案分别生成日前出力场景，其生成场景集置信区间计算结果如附录 A 表 A5 所示。

从表 A5 中可以看出，对于两个地区的风电场三种方案均有着较高的覆盖率，这是由于三种方案均为数据驱动的方法，对于数据的分布有着一定的挖掘能力。但是，方案一较其余两种方法所生成的场景更为保守，其功率区间宽度更宽。而方案三在不同数据集下所生成的场景在相近覆盖率下均有着较窄的功率区间宽度。由此可见，本文所提出的方法在不同数据集下生成日前场景也有着较高的准确性。

同时为了验证本文方法在不同数据集下的随机性，统计三种方案在不同风电场下的欧式距离平均值，计算结果如附录 A 表 A6 所示。

从表 A6 中可以看出，方案三在不同地区的风电场下的欧式距离平均值均较小。因此，由本文所提出的方法在不同数据集下生成日前场景有着较低的随机性。

综上所述，方案一、方案二与方案三在描述新能源出力不确定性方面均有着较好的效果，但方案三所生成的场景集准确性更好，随机性更低，所生成的场景更符合实际情况。方案一所刻画的出力不确定性区间更为保守，随机性较大。方案二较方案一生成效果有所提升，但由于未引入新能源预测信息及使用线性噪声表添加噪声，所生成场景集功率区间宽度相对较宽，欧式距离平均值相对较大。通过上述的对比结果，可以看出本文所提出的基于改

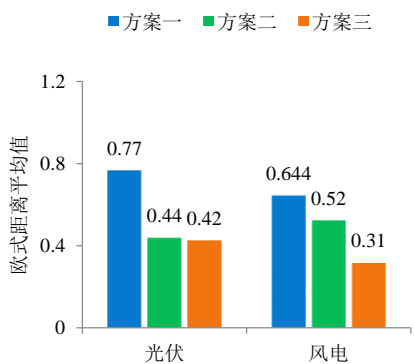

进条件生成扩散模型的场景生成方法能够较为准确的描述新能源出力的不确定性。

## 4 结论

针对基于现有场景生成方法存在适应性和可解释性较差的问题，本文提出了基于改进条件生成扩散模型的新能源日前场景生成方法。该方法基于马尔可夫链与变分推理的理论框架，其加噪以及去噪过程的每一步均有着马尔可夫性质，从而可以了解模型在每一步生成的结果，且通过变分下界最大化来训练模型。最后以算例中以真实的风光出力数据验证了本文方法的有效性，具体结论如下：

1)ICGDM 能够以数据驱动的方式自适应学习风光出力的特征及变化规律，从而有效生成符合新能源变化趋势的场景。

2)将本文方法与原始生成扩散模型以及WGAN 场景生成方法的生成效果对比可见。本文方法所生成的场景集在同一置信度下，对实际场景均有着较高的覆盖率和较窄的功率区间宽度，随机性较小。此外，在不同地区的风光出力场景生成中也均有着较高的准确性和较小的欧氏距离平均值。由此表明，本文方法能够较为准确描述新能源日前出力不确定性。

3)本文方法不涉及显式特征分析，具有一定的适应性，为新能源日前场景生成提供了一种新方法。在未来的研究中，应将风光之间的互补特性及多风场之间的关联特性纳入考虑范围。

附录见本刊网络版(http://www.dwjs.com.cn/CN/1000-3673/current.shtml)。

附录 A

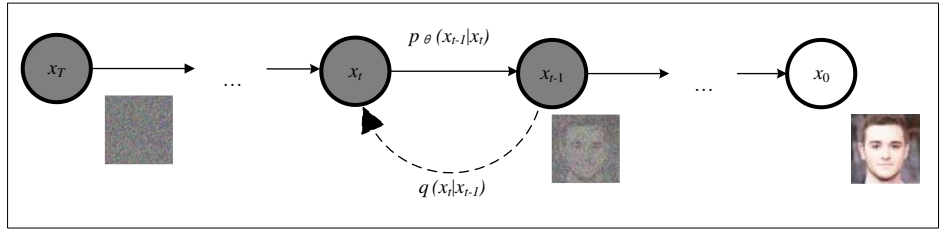

图 A1 生成扩散模型理论框图

Fig.A1 Theoretical framework diagram of the generative diffusion model

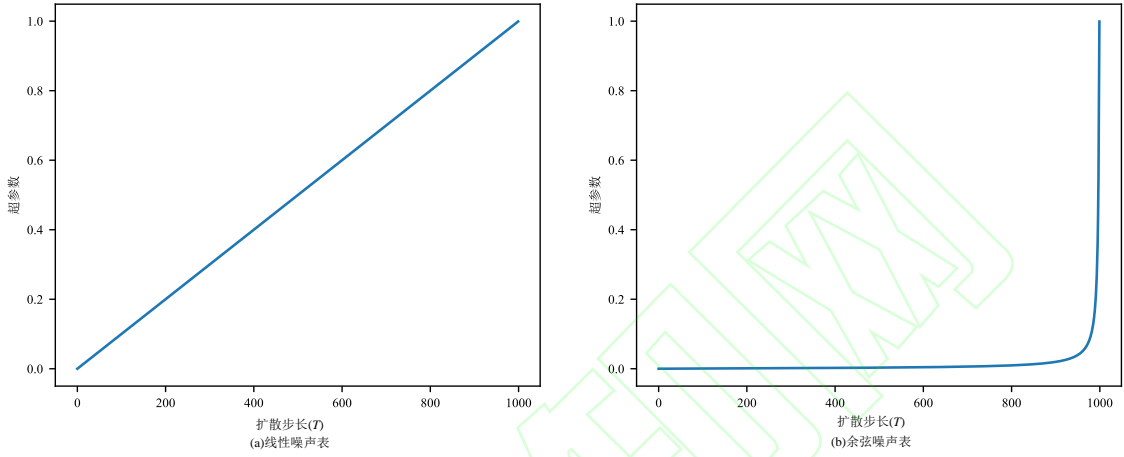

图 A2 噪声表

Fig.A2 Noise schedule

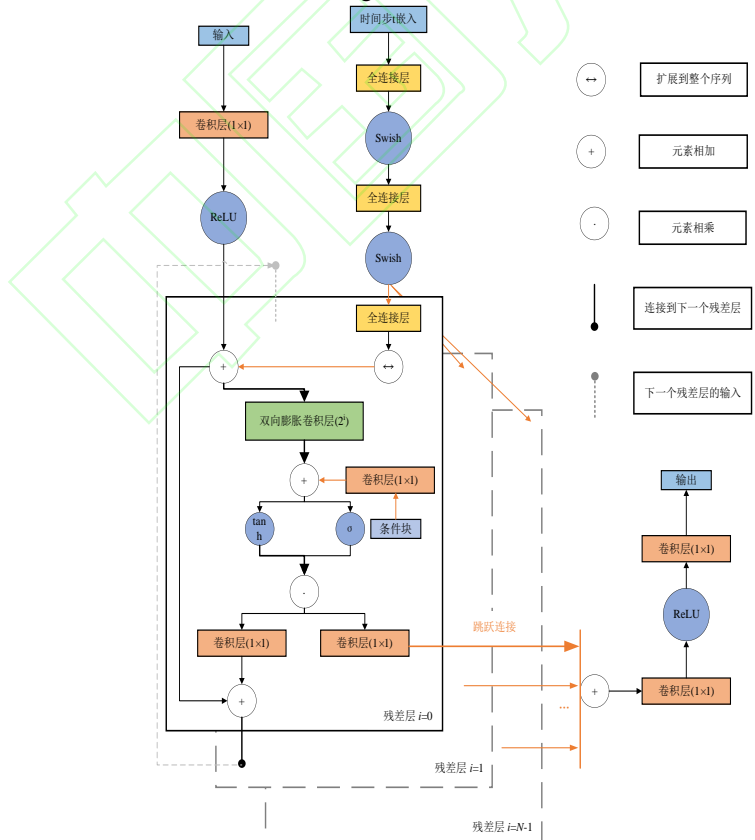

图 A3 神经网络结构图

Fig.A3 Neural Network Architecture Diagram

表 A1 神经网络参数

Tab.A1 Neural network parameter

| 参数名称 | 参数大小 |
|---|---|
| 时间嵌入维度 | 16 |
| 残差块数量 | 8 |
| 残差通道数量 | 8 |
| 膨胀周期长度 | 2 |
| 残差块中隐藏层大小 | 64 |

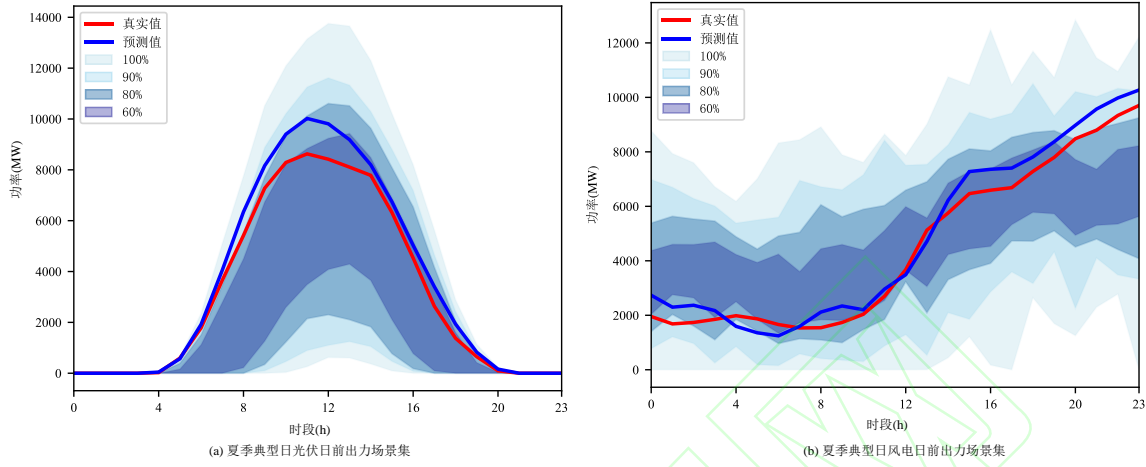

(a) 夏季典型日光伏日前出力场景集　　　(b) 夏季典型日风电日前出力场景集

图 A4　夏季典型日方案一生成场景集的置信区间

Fig.A4 Confidence Interval of the Generated Scenario Set for Scheme 1 on a Typical Summer Day

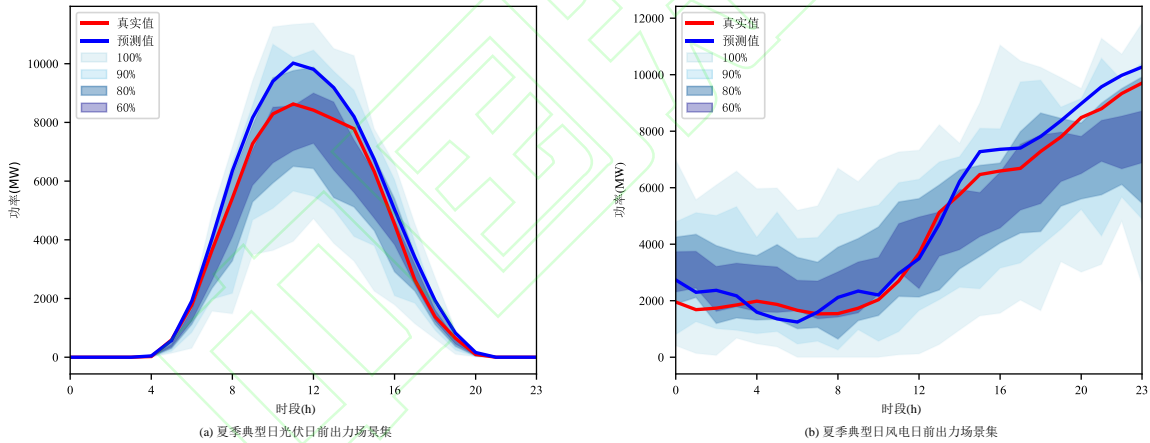

(a) 夏季典型日光伏日前出力场景集　　　(b) 夏季典型日风电日前出力场景集

图 A5　夏季典型日方案二生成场景集的置信区间

Fig.A5 Confidence Interval of the Generated Scenario Set for Scheme 2 on a Typical Summer Day

表 A2　夏季典型日不同方案生成场景集的置信区间

Tab.A2 Confidence Intervals of the Generated Scenario Sets for Different Schemes on a Typical Summer Day

|  | 置信度 | 方案一 | | 方案二 | | 方案三 | |
|---|---|---|---|---|---|---|---|
|  |  | 覆盖率/% | 功率区间/MW | 覆盖率/% | 功率区间/MW | 覆盖率/% | 功率区间/MW |
| 光伏 | 100% | 100 | 8853.339 | 100 | 6858.181 | 100 | **4765.854** |
|  | 90% | 100 | 6272.418 | 100 | 4651.974 | 100 | **3313.958** |
|  | 80% | 83.333 | 4096.569 | 91.667 | 3059.899 | 95.833 | **2197.816** |
|  | 60% | 45.833 | 2439.185 | 58.333 | 1696.318 | 79.167 | **1213.733** |
| 风电 | 100% | 100 | 4904.491 | 95.833 | 2885.174 | 100 | **2166.651** |
|  | 90% | 87.5 | 3964.571 | 95.833 | 2008.368 | 95.833 | **1514.184** |
|  | 80% | 75 | 3311.367 | 91.667 | 1352.188 | 91.667 | **1071.189** |
|  | 60% | 58.333 | 2272.750 | 58.333 | 708.851 | 87.5 | **586.658** |

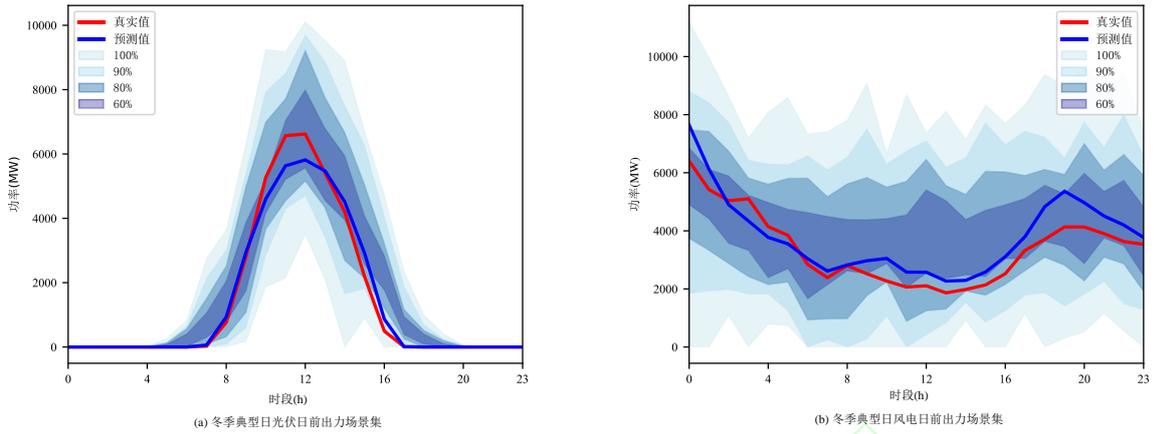

图 A6 冬季典型日方案一生成场景集的置信区间

Fig.A6 Confidence Interval of the Generated Scenario Set for Scheme 1 on a Typical Winter Day

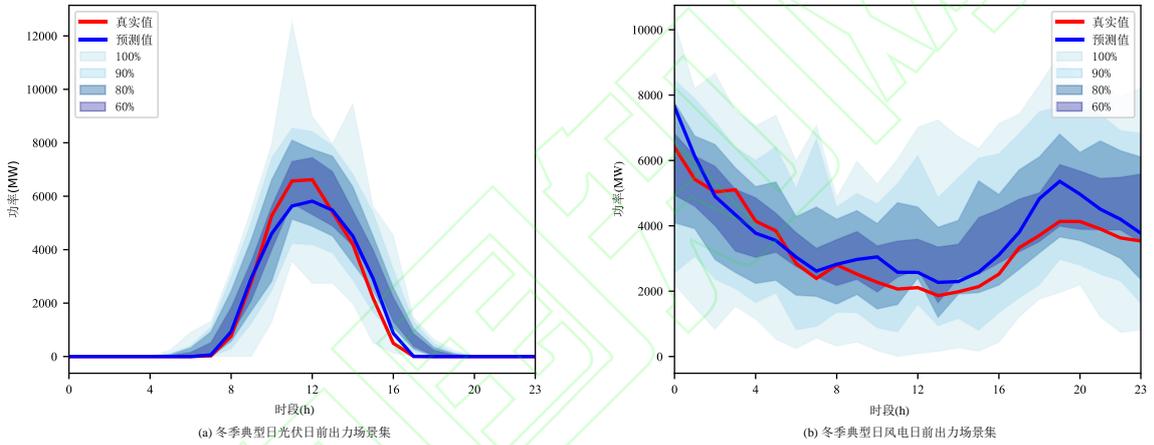

图 A7 冬季典型日方案二生成场景集的置信区间

Fig.A7 Confidence Interval of the Generated Scenario Set for Scheme 2 on a Typical Winter Day

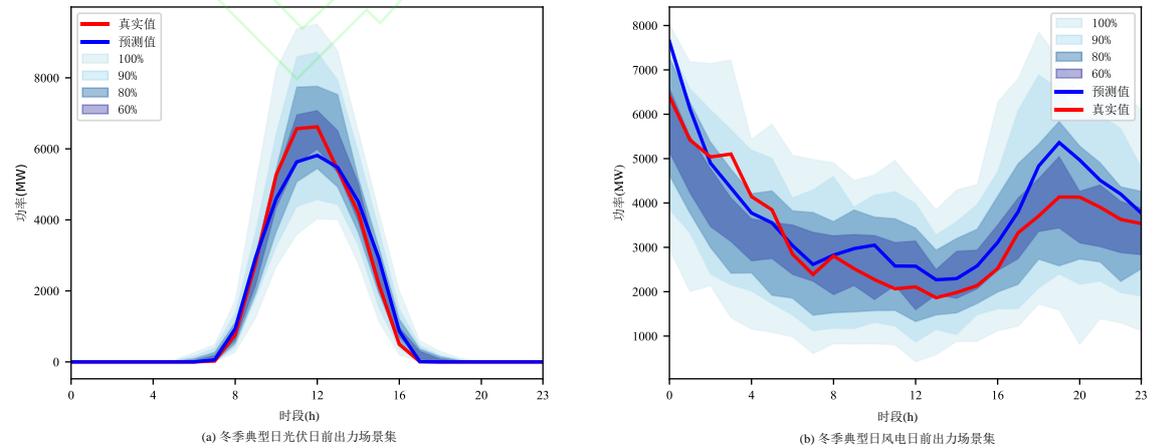

图 A8 冬季典型日方案三生成场景集的置信区间

Fig.A8 Confidence Interval of the Generated Scenario Set for Scheme 3 on a Typical Winter Day

表 A3 冬季典型日不同方案生成场景集的置信区间

Tab.A3 Confidence Intervals of the Generated Scenario Sets for Different Schemes on a Typical Winter Day

| | 置信度 | 方案一 | | 方案二 | | 方案三 | |
|---|---|---|---|---|---|---|---|
| | | 覆盖率/% | 功率区间/MW | 覆盖率/% | 功率区间/MW | 覆盖率/% | 功率区间/MW |
| 光伏 | 100% | 100 | 3568.139 | 100 | 2411.989 | 100 | **1560.636** |
| | 90% | 95.833 | 2737.603 | 91.667 | 1571.203 | 100 | **1053.025** |
| | 80% | 87.5 | 1846.632 | 87.5 | 1088.671 | 91.667 | **645.134** |
| | 60% | 58.33 | 1106.932 | 75 | 610.727 | 70.833 | **337.657** |
| 风电 | 100% | 100 | 8100.673 | 100 | 6510.519 | 100 | **4496.629** |
| | 90% | 100 | 5814.273 | 100 | 4649.081 | 100 | **3200.673** |
| | 80% | 100 | 3907.079 | 95.833 | 2950.053 | 87.5 | **2150.994** |
| | 60% | 75 | 2182.439 | 70.833 | 1529.542 | 79.167 | **1291.913** |

表 A4 冬季典型日三种方案欧式距离平均值对比

Tab.A4 Comparison of the average Euclidean distance among the three typical summer day plans

| 随机性 | 方案一 | 方案二 | 方案三 |
|---|---|---|---|
| 光伏 | 0.933 | 0.589 | **0.459** |
| 风电 | 0.936 | 0.531 | **0.424** |

表 A5 不同地区下场景生成置信区间计算结果

Tab.A5 Scene generates confidence interval calculation results

| | 置信度 | 方案一 | | 方案二 | | 方案三 | |
|---|---|---|---|---|---|---|---|
| | | 覆盖率/% | 功率区间/MW | 覆盖率/% | 功率区间/MW | 覆盖率/% | 功率区间/MW |
| 瓦隆地区 | 100% | 100 | 6371.859 | 100 | 4618.632 | 100 | **3146.734** |
| | 90% | 100 | 4716.378 | 100 | 3045.814 | 100 | **2064.654** |
| | 80% | 100 | 3133.313 | 100 | 2007.088 | 100 | **1342.246** |
| | 60% | 91.667 | 1755.365 | 95.833 | 1034.368 | 95.833 | **736.505** |
| 弗兰德伦地区 | 100% | 100 | 2066.747 | 100 | 1725.961 | 100 | **1375.448** |
| | 90% | 100 | 1570.581 | 100 | 1159.003 | 100 | **949.789** |
| | 80% | 83.333 | 983.571 | 91.667 | 732.463 | 91.667 | **631.929** |
| | 60% | 75 | 545.673 | 66.667 | 405.082 | 83.333 | **357.32** |

表 A6 不同风电场的欧式距离平均值

Tab.A6 Average European distance of different wind farms

| 随机性 | 方案一 | 方案二 | 方案三 |
|---|---|---|---|
| 风电场 1 | 1.038 | 0.962 | **0.766** |
| 风电场 2 | 0.537 | 0.466 | **0.45** |

# The Day-ahead Scenario Generation Method for New Energy Based on an Improved Conditional Generative Diffusion Model

WANG Changgang[1,2], LIU Wei[2], CAO Yu[1,2], LIANG Dong[3], LI Yang[1,2], MO Jingshan[1,2]

(1. Key Laboratory of Modern Power System Simulation and Control and Renewable Energy Technology, Ministry of Education, Northeast Electric Power University; 2. School of Electrical Engineering, Northeast Electric Power University; 3. Jilin Power Supply Company Substation Secondary Maintenance Center)

As the proportion of renewable energy in power generation continues to increase, accurately generating renewable energy output scenarios is crucial for effective day-ahead power system scheduling. While scenario analysis methods are commonly used to describe the variability of wind and solar power output, existing methods often suffer from poor adaptability and limited interpretability.

In response to these challenges, this paper proposes an improved conditional generative diffusion model for generating renewable energy scenarios. The method takes historical measured data and day-ahead forecast data as real samples and conditional inputs, respectively. Initially, noise is progressively added to the original samples during the diffusion process, transforming the historical data into pure noise. Subsequently, the model employs a denoising process guided by conditional information to gradually remove the noise and restore the original data, ultimately generating data that meets the specified conditional distribution.

After incorporating the conditional information, the neural network's target becomes the noise that satisfies the conditional distribution, denoted as $\varepsilon_\theta(x_t,t,c)$. At this point, the loss function is expressed as:

$$L_t = E_{x_0,\varepsilon}\left[\left\|\varepsilon(x_t,t) - \varepsilon_\theta(x_t,t,c)\right\|^2\right] \quad (31)$$

Where $\|\cdot\|$ represents the L2 norm; $\varepsilon_\theta$ denotes the predicted noise.

Overall, this method utilizes a theoretical framework based on Markov chains and variational inference. In the forward process, a defined noise schedule allows for the explicit determination of the noise added at each step, thereby capturing the uncertainty generated by the model at each stage. In the reverse process, the denoising at each step is based on the result of the previous step, with a well-defined probability distribution for each, enhancing the transparency of the entire generation process. Additionally, training the model by maximizing the variational lower bound further improves interpretability. Moreover, this paper replaces the linear noise schedule with a cosine noise schedule, enhancing the model's adaptability. A framework diagram of this method is shown below.

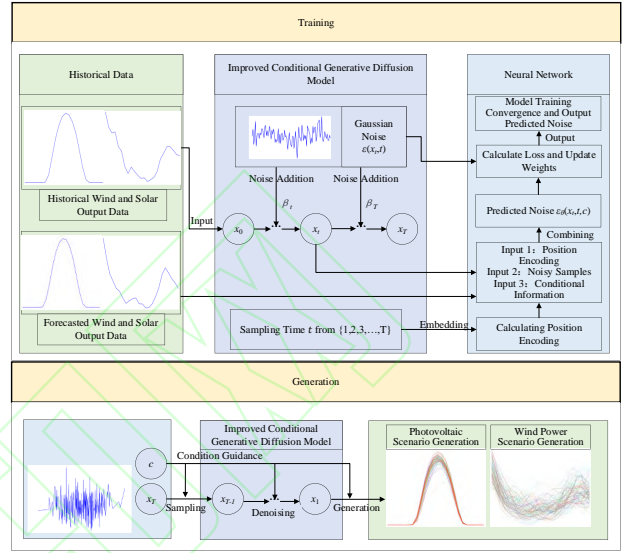

**Fig.1 New energy scenario generation method framework based on improved condition generation diffusion model**

A case analysis was conducted using Belgium's day-ahead forecast and actual data. The results indicate that the output trends for photovoltaic and wind power scenarios generated by this method closely align with the forecasted values, with the real values being well captured within the generated scenario set and showing no significant random fluctuations. The method's effectiveness was further validated using autocorrelation coefficients.

Finally, a comparison with the original generative diffusion model and the WGAN scenario generation method revealed that the scenario set generated by this method offers higher coverage of actual scenarios and narrower power interval widths at the same confidence level, with less randomness. Additionally, the method demonstrated high accuracy and lower mean Euclidean distances in generating wind and solar output scenarios across different regions. This method proves to be effective in accurately generating day-ahead new energy scenarios.